# Collaborative Representation based Classification for Face Recognition

Lei Zhang[a,1], Meng Yang[a], Xiangchu Feng[b], Yi Ma[c], and David Zhang[a]

[a] Dept. of Computing, The Hong Kong Polytechnic University, Hong Kong, China
[b] School of Applied Mathematics, Xidian University, Xi'an, China
[c] Principal Researcher, Microsoft Research Asia, Beijing, China

**Abstract:** By coding a query sample as a sparse linear combination of all training samples and then classifying it by evaluating which class leads to the minimal coding residual, sparse representation based classification (SRC) leads to interesting results for robust face recognition. It is widely believed that the $l_1$-norm sparsity constraint on coding coefficients plays a key role in the success of SRC, while its use of all training samples to collaboratively represent the query sample is rather ignored. In this paper we discuss how SRC works, and show that the collaborative representation mechanism used in SRC is much more crucial to its success of face classification. The SRC is a special case of collaborative representation based classification (CRC), which has various instantiations by applying different norms to the coding residual and coding coefficient. More specifically, the $l_1$ or $l_2$ norm characterization of coding residual is related to the robustness of CRC to outlier facial pixels, while the $l_1$ or $l_2$ norm characterization of coding coefficient is related to the degree of discrimination of facial features. Extensive experiments were conducted to verify the face recognition performance of CRC with different instantiations.

---

[1] Corresponding author. Email: cslzhang@comp.polyu.edu.hk.

# 1. Introduction

It has been found that natural images can be coded by a small number of structural primitives which are qualitatively similar in form to simple cell receptive fields [1-2]. In the past decade, the sparse coding or sparse representation methods have been rapidly developed. Sparse representation codes a signal $y$ over an over-complete dictionary $\Phi$ such that $y \approx \Phi \alpha$ and $\alpha$ is a sparse vector. The sparsity of $\alpha$ is often characterized by $l_1$-norm, leading to the sparse coding model: $\min_\alpha \|\alpha\|_1 \ s.t. \ \|y - \Phi\alpha\|_2 \leq \varepsilon$, where $\varepsilon$ is a small constant [7-9]. The successful applications of sparse representation include image restoration [3-6], compressive sensing [10], morphological component analysis [11], and super-resolution [12-13], etc.

The great success of sparse representation in image reconstruction triggers the research on sparse representation based pattern classification. The basic idea is to code the test sample over a dictionary with sparsity constraint, and then classify it based on the coding vector. Huang and Aviyente [14] sparsely coded a signal over a set of predefined redundant bases and used the coding vector as features for classification. Rodriguez and Sapiro [15] learned a discriminative dictionary to code the image for classification. In [16], Wright *et al.* proposed a very interesting method, namely sparse representation based classification (SRC), for face recognition (FR). Denote by $X_i \in \Re^{m \times n}$ the set of training samples from class $i$ (each column of $X_i$ is a sample). Suppose that we have $K$ classes of subjects, and let $X = [X_1, X_2, \ldots, X_K]$. For a query face image $y \in \Re^m$, it is coded over $X$ as $y \approx X\alpha$, where $\alpha = [\alpha_1; \ldots; \alpha_i; \ldots; \alpha_K]$ and $\alpha_i$ is the sub-vector associated with $X_i$. If $y$ is from class $i$, usually $y \approx X_i \alpha_i$ holds well, implying that most coefficients in $\alpha_k$, $k \neq i$, are small and only $\alpha_i$ has significant values. That is, the sparse non-zero entries in $\alpha$ can encode the identity of $y$. The procedures of standard SRC are summarized in Table 1.

**Table 1**: The standard SRC Algorithm

1. Normalize the columns of $X$ to have unit $l_2$-norm.
2. Code $y$ over $X$ via $l_1$-norm minimization

$$\hat{\alpha} = \arg\min_\alpha \left\{ \|y - X\alpha\|_2^2 + \lambda \|\alpha\|_1 \right\} \quad (1)$$

where $\lambda$ is a positive scalar.
3. Compute the residuals

$$r_i = \|y - X_i \hat{\alpha}_i\|_2 \quad (2)$$

where $\hat{\alpha}_i$ is the coefficient vector associated with class $i$.
4. Output the identity of $y$ as

$$\text{identity}(y) = \arg\min_i \{r_i\} \quad (3)$$

We denote by S-SRC the standard SRC scheme described in Table 1. When $y$ is occluded or corrupted, Wright *et al.* [16] used the identity matrix $I$ as an additional dictionary to code the outlier pixels, i.e., $\hat{\alpha} = \arg\min_{\alpha,\beta} \{\|y - [X, I][\alpha; \beta]\|_2^2 + \lambda \|[\alpha; \beta]\|_1\}$. It can be seen that this coding model is equivalent to $\hat{\alpha} = \arg\min_{\alpha} \{\|y - X\alpha\|_1 + \lambda \|\alpha\|_1\}$; that is, the coding residual is also characterized by $l_1$-norm to achieve robustness to outliers. We denote by R-SRC this robust SRC scheme to deal with occlusions and corruption.

SRC (including S-SRC and R-SRC) shows very interesting robust FR performance, and it boosts much the research of sparsity based pattern classification. Inspired by SRC, Gao *et al.* [17] proposed the kernel sparse representation for FR, while Yang and Zhang [18] learned a Gabor occlusion dictionary to reduce significantly the computational cost when dealing with face occlusion. Cheng *et al.* [19] constructed the $l_1$-graph for image classification, while Qiao *et al.* [20] learned a subspace to preserve the $l_1$-graph for FR. In [21], Yang *et al.* combined sparse coding with linear pyramid matching for image classification. In SRC, it is assumed that face images are aligned, and methods have also been proposed to solve the misalignment or pose change problem. For example, the method in [22] is invariant to image-plane transformation, and the method in [23] was designed to deal with misalignment and illumination variations. Peng *et al.* [24] used low-rank decomposition to align a batch of linearly correlated images with gross corruption. In addition, dictionary learning methods [25-28] were also developed to enhance SRC based pattern classification.

The $l_1$-minimization required in sparsity based pattern classification can be time consuming. Many fast algorithms have been proposed to speed up the $l_1$-minimization process [29-36]. As reviewed in [31], there are five representative fast $l_1$-norm minimization approaches, namely, Gradient Projection, Homotopy, Iterative Shrinkage-Thresholding, Proximal Gradient, and Augmented Lagrange Multiplier (ALM). It was indicated in [31] that for noisy data, the first order $l_1$-minimization techniques (e.g., SpaRSA [32], FISTA [33], and ALM [34]) are more efficient, while in the application of FR, Homotopy [35], ALM and $l_1$_ls [36] are better for their good accuracy and fast speed.

Though SRC has been widely studied in the FR community, its working mechanism is not fully revealed yet. The role of $l_1$-sparsity is often emphasized, and many works aim to improve the $l_1$-regularization on coding vector $\alpha$. For examples, Liu *et al.* [37] added a nonnegative constraint to $\alpha$; Gao *et al.* [38] introduced a Laplacian term of $\alpha$ in sparse coding; Yuan and Yan [39] used joint sparse representation to code multiple types of image features; and Elhamifar and Vidal [40] used structured sparse representation for robust classification. All these works stress the role of $l_1$-sparsity of $\alpha$ in classification. However, the use of training samples from all classes to represent the query sample $y$ in SRC is rather ignored. Some recent works [41-42] have questioned the role of sparsity in pattern classification. Berkes *et al.* [41] argued that there is no clear evidence for active sparsification in the visual cortex. In our previous work [42], it is shown that the use of collaborative representation is more crucial than the $l_1$-sparsity of $\alpha$ to FR, and the $l_2$-norm regularization on $\alpha$ can do a similar job to $l_1$-norm regularization but with much less computational cost.

The SRC classifier has a close relationship to the nearest classifiers, including the nearest neighbor (NN), nearest feature line (NFL) [43], nearest feature plane (NFP) [44], local subspace (LS) [45] and nearest subspace (NS) [44][46-48]. The NN, NFL and NSP classifiers use one, two and three training samples of each class, respectively, to represent the query image $y$, while the LS and NS classifiers represent $y$ by all the training samples of each class. Like these nearest classifiers, SRC also represents $y$ as the linear combination of training samples; however, one critical difference is that SRC represents $y$ by the training samples from all classes, while the nearest classifiers represent $y$ by each individual class. The use of all classes to collaboratively represent $y$ alleviates much the small-sample-size problem in FR, especially when the number of training samples per class is small.

In this paper, we discuss in detail the collaborative representation nature of SRC. Compare with our previous work in [42], in this paper we present a more general model, namely collaborative representation based classification (CRC), make deeper analysis on the $l_1/l_2$-norm regularization of coding coefficients, and conduct more experiments. By using $l_1$-norm or $l_2$-norm to characterize the coding vector $\alpha$ and the coding residual $e=y-X\alpha$, we can have different instantiations of CRC, while S-SRC and R-SRC are special cases of CRC. The $l_1$- or $l_2$-norm characterization of $e$ is related to the robustness of CRC to outlier pixels, while the $l_1$- or $l_2$-norm characterization of $\alpha$ is related to the discrimination of facial feature $y$. When the face image is not occluded/corrupted, $l_2$-norm is good enough to model $e$; when the face image is occluded/corrupted, $l_1$-norm is more robust to model $e$. The discrimination of facial feature $y$ is often related to its dimensionality. If the dimensionality and the discrimination of $y$ is high, the coding coefficients $\alpha$ will be naturally sparse and concentrate on the samples whose class label is the same as $y$, no matter $l_1$- or $l_2$-norm is used to regularize $\alpha$. When the dimensionality of $y$ is too low, often the discrimination power of $y$ will be much reduced, and the distribution of $\alpha$ will be less sparse since some big coefficients can be generated and assigned to samples whose class labels are different from *y*. In this case, the $l_1$-norm regularization on $\alpha$ will enforce $\alpha$ to be sparse, and consequently enhance its discrimination power.

The rest of this paper is organized as follows. Section 2 discusses the role of sparsity in face representation and classification. Section 3 discusses in detail the CRC scheme. Section 4 conducts extensive experiments to demonstrate the performance of CRC, and Section 5 concludes the paper.

## 2. The Role of Sparsity in Representation based Face Recognition

As shown in Table 1, there are two key points in SRC [16]: (i) the coding vector $\alpha$ is enforced to be sparse and (ii) the coding of query sample *y* is performed over the whole dataset *X* instead of each subset $X_i$. It was stated in [16] that the sparsest (or the most compact) representation of *y* over *X* is naturally discriminative and can indicate the identity of *y*. The SRC is a generalization and significant extension of classical nearest classifiers such as NN and NS by representing *y* collaboratively across classes. But there are some issues not very clear yet: is it the sparsity constraint on $\alpha$ that makes the representation more discriminative, and must we use $l_1$-sparsity to this end?

Denote by $\Phi \in \Re^{m \times n}$ a dictionary of bases (atoms). If $\Phi$ is complete, then any signal $x \in \Re^m$ can be accurately represented as the linear combination of the atoms in $\Phi$. If $\Phi$ is orthogonal and complete, however, often we need to use many atoms from $\Phi$ to faithfully represent *x*. If we want to use less atoms to represent *x*, we must relax the orthogonality requirement on $\Phi$. In other words, more atoms should be involved in $\Phi$ so that we have more choices to represent *x*, leading to an over-complete and redundant dictionary $\Phi$ but a sparser representation of *x*. The great success of sparse representation in image restoration [3-6] validates that a redundant dictionary can have more powerful capability to reconstruct the signal.

In the scenario of FR, each class of face images often lies in a small subspace of $\Re^m$. That is, the *m*-dimensional face image *x* can be characterized by a code of much lower dimensionality. Let's take the training samples of class *i*, i.e., $X_i$, as the dictionary of this class. In practice the atoms (i.e., the training samples) of $X_i$ will be correlated. Assume that we have enough training samples of each class and all the face images of class *i* can be faithfully represented by $X_i$, then $X_i$ can be viewed as an over-complete dictionary[2] because of the correlation of training samples of class $i$. Therefore, we can conclude that a test sample $y$ of class $i$ can be sparsely represented by dictionary $X_i$.

Another important fact in FR is that human faces are all somewhat similar, and some subjects may have very similar face images. Dictionary $X_i$ of class $i$ and dictionary $X_j$ of class $j$ are not incoherent; instead, they can be highly correlated. Using the NS classifier, for a query sample $y$ from class $i$, we can find (by least square method) a coding vector $\alpha_i$ such that $\alpha_i = \arg\min_\alpha \|y - X_i\alpha\|_2^2$. Let $r_i = y - X_i\alpha_i$. Similarly, if we represent $y$ by class $j$, there is $\alpha_j = \arg\min_\alpha \|y - X_j\alpha\|_2^2$ and we let $r_j = y - X_j\alpha_j$. For the convenience of discussion, we assume that $X_i$ and $X_j$ have the same number of atoms, i.e., $X_i, X_j \in \Re^{m \times n}$. Let $X_j = X_i + \Delta$. When $X_i$ and $X_j$ are very similar, $\Delta$ can be very small such that $\xi = \frac{\|\Delta\|_F}{\|X_i\|_F} \leq \frac{\sigma_n(X_i)}{\sigma_1(X_i)}$, where $\sigma_1(X_i)$ and $\sigma_n(X_i)$ are the largest and smallest eigenvalues of $X_i$, respectively. Then we can have the following relationship between $r_i$ and $r_j$ (Theorem 5.3.1, page 242, [49]):

$$\frac{\|r_j - r_i\|_2}{\|y\|_2} \leq \xi\left(1 + \kappa_2(X_i)\right)\min\{1, m-n\} + O\left(\xi^2\right) \quad (4)$$

where $\kappa_2(X_i)$ is the $l_2$-norm conditional number of $X_i$. From Eq. (4), we can see that if $\Delta$ is very small, the distance between $r_i$ and $r_j$ will also be very small. This makes the classification very unstable because some small disturbance (e.g., noise or small deformation) can make $\|r_j\|_2 < \|r_i\|_2$, leading to a wrong classification.

The above problem can be alleviated by regularization on $\alpha_i$ and $\alpha_j$. The reason is very intuitive. Take the $l_0$-norm sparsity regularization as an example, if $y$ is from class $i$, it is more likely that we can use only a few samples, e.g., 5 or 6 samples, in $X_i$ to represent $y$ with a good accuracy. In contrast, we may need more samples, e.g., 8 or 9 samples, in $X_j$ to represent $y$ with nearly the same representation accuracy. With the sparsity constraint or other regularizer, the representation error of $y$ by $X_i$ will be visibly lower than that by $X_j$, making the classification of $y$ easier. Here let's consider three regularizers: the sparse regularizers by $l_0$-norm and $l_1$-norm, and the least square regularizer by $l_2$-norm.

---

[2]More strictly speaking, it should be the dimensionality reduced dictionary of $X_i$ that is over-complete. For the convenience of expression, we simply use $X_i$ in the development.

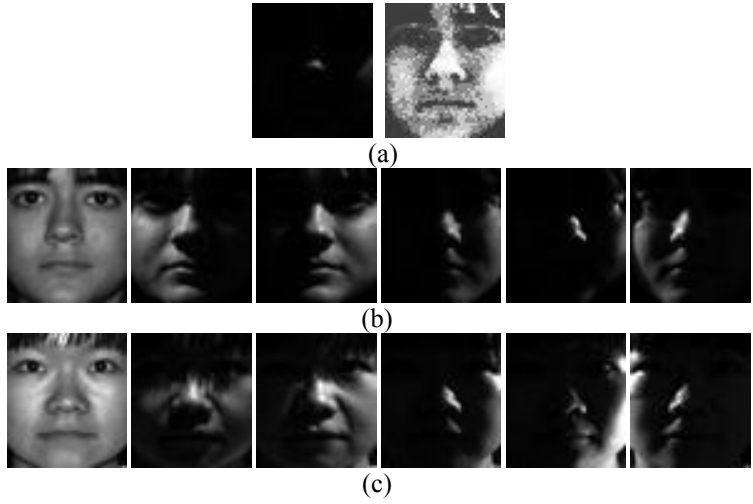

**Figure 1:** An example of class-specific face representation. (a) The query face image (left: original image; right: the one after histogram equalization for better visualization); (b) some training samples from the class of the query image; (c) some training samples from another class.

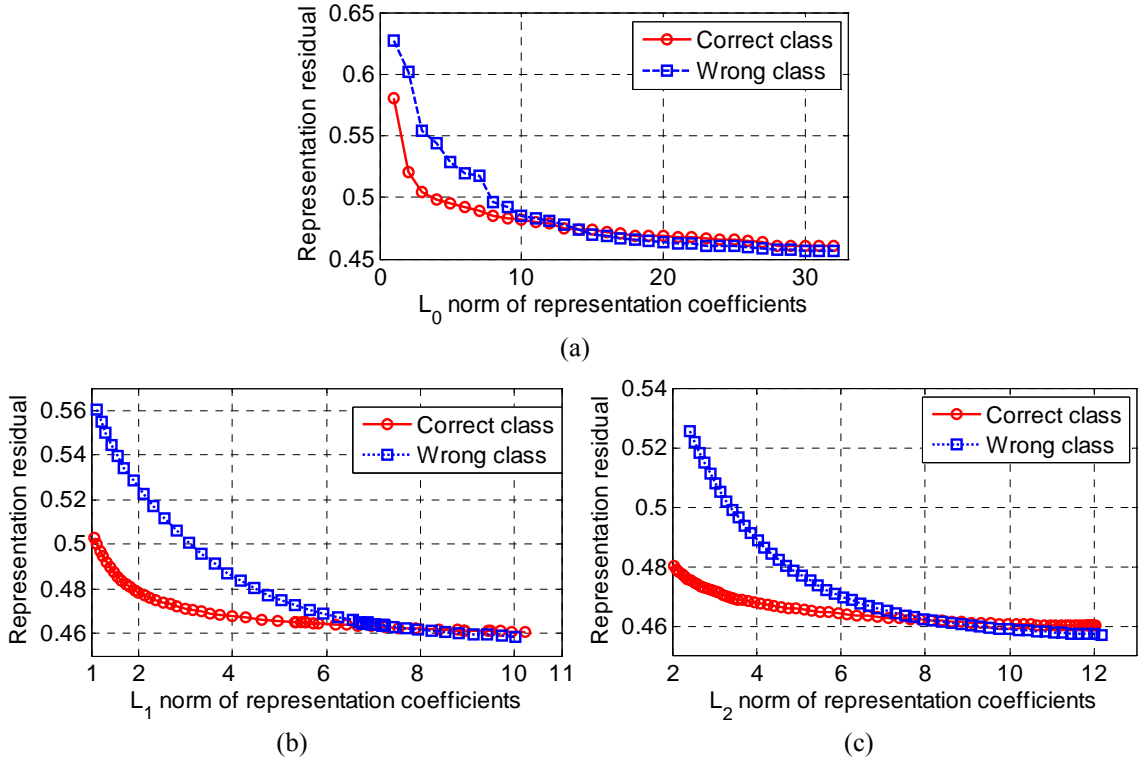

**Figure 2:** The curve of representation residual versus the $l_p$-norm of representation coefficients. (a) $p=0$, (b) $p=1$, and (c) $p=2$.

By $l_p$-regularization, $p = 0$, 1, or 2, the representation of $y$ over dictionary $\Phi$ can be formulated as

$$\hat{\alpha} = \arg\min_{\alpha} \|y - \Phi\alpha\|_2^2 \quad \text{s.t.} \quad \|\alpha\|_{l_p} \leq \varepsilon \tag{5}$$

where $\varepsilon$ is a small positive number. Let $r = \|y - \Phi\hat{\alpha}\|_2$. We could plot the curve of "$r$ vs. $\varepsilon$" to illustrate how regularization improves discrimination. Fig. 1(a) shows a test face image of class 32 in the Extended Yale B database [46][50]. Some training samples of this class are shown in Fig. 1(b), while some training samples of class 5, which is similar to class 32, are shown in Fig. 1(c). We use the training samples of the two classes as dictionaries to represent, respectively, the query sample in Fig. 1(a) via Eq. (5). The "$r$ vs. $\varepsilon$" curves for $p = 0$, 1, and 2 are drawn in Figs. 2 (a)~(c), respectively. For the $l_0$-norm regularization, we used the Orthogonal Matching Pursuit (OMP) algorithm [51] to solve Eq. (5); for the $l_1$-norm regularization, we used the $l_1\_ls$ algorithm [36]; while for $l_2$-norm regularization, the regularized least square was used to get an analytical solution to Eq. (5).

From Fig. 2(a), one can see that when only a few training samples (e.g., less than 3 samples) are used to represent the query sample, both the two classes have big representation error. In practice, the system will consider this sample as an imposter and directly reject it. When more and more training samples are involved, the representation residual $r$ decreases. However, the ability of $r$ to discriminate the two classes will also reduce if too many samples (e.g., more than 10 samples) are used to represent the query sample. This is because the two classes are similar so that the dictionary of one class can represent the samples of another class if enough training samples are available (i.e., the dictionary is nearly over-complete). With these observations, one can conclude that a query sample should be classified to the class which could faithfully represent it using fewer samples. In other words, the $l_0$-norm sparse regularization on $\alpha$ can improve the discrimination of representation based classification.

Now the question is: can the weaker $l_1$-regularization, and even the non-sparse $l_2$-regularization, do a similar job? Fig. 2(b) and Fig. 2(c) give the answer. We can see that when $\varepsilon$ is big ($\varepsilon > 8$), which means that the regularization is weak, both the two classes have very low reconstruction residual, making the classification very unstable. By setting a smaller $\varepsilon$, the $l_1$-norm or $l_2$-norm regularization on $\alpha$ will lead to a discriminative reconstruction residual, by which the query sample can be correctly classified. From this example, one can see that the non-sparse $l_2$-norm regularization can play a similar role to the sparse $l_0$-norm or $l_1$-norm regularization in enhancing the discrimination of representation.

**Remark (regularized nearest subspace, RNS):** The above observations imply a regularized nearest subspace (RNS) scheme for FR when the number of training samples of each class is big. That is, we can represent the query sample $y$ class by class, and classify it based on the representation residual and regularization strength. Since $l_0$-norm minimization is combinatorial and NP-hard, it is more practical to use $l_1$-norm or $l_2$-norm to regularize the representation coefficients. Using the Lagrangian formulation, we have the objective function of RNS-$l_p$ as:

$$\hat{\boldsymbol{\alpha}} = \arg\min_{\boldsymbol{\alpha}} \left\{ \|\boldsymbol{y} - \boldsymbol{\Phi}\boldsymbol{\alpha}\|_2^2 + \lambda \|\boldsymbol{\alpha}\|_{l_p} \right\} \quad (6)$$

where $p = 1$ or $2$ and $\lambda$ is a positive constant. For each class $X_i$, we could obtain its representation vector $\hat{\boldsymbol{\alpha}}_i$ of $\boldsymbol{y}$ by taking $\boldsymbol{\Phi}$ as $X_i$ in Eq. (6). Denote by $r_i = \|\boldsymbol{y} - X_i\hat{\boldsymbol{\alpha}}_i\|_2^2 + \lambda \|\hat{\boldsymbol{\alpha}}_i\|_{l_p}$, and we can then classify $\boldsymbol{y}$ by $\text{identity}(\boldsymbol{y}) = \arg\min_i \{r_i\}$.

## 3. Collaborative Representation based Classification (CRC)

In Section 2, it is assumed that each class has enough training samples. Unfortunately, FR is a typical small-sample-size problem, and $X_i$ is under-complete in general. If we use $X_i$ to represent $\boldsymbol{y}$, the representation residual $r_i$ can be big even when $\boldsymbol{y}$ is from class $i$. This problem can be overcome if more samples of class $i$ can be used to represent $\boldsymbol{y}$, yet the problem is how to find the additional samples. Fortunately, one fact in FR is that the face images of different people share certain similarities, and some subjects, say subject $i$ and subject $j$, can be very similar so that the samples from class $j$ can be used to represent the test sample of class $i$. In other words, one class can borrow samples from other classes in order to faithfully represent the query sample. Such a strategy is very similar to the nonlocal technique widely used in image restoration [52-54], where for a given local patch the many similar patches to it are collected throughout the image to help the reconstruction of the given patch. In the scenario of FR, for each class we may consider the samples from similar classes as the "nonlocal samples" and use them to better represent the query sample.

However, such a "nonlocal" strategy has some problems to implement under the scenario of FR. First, how to find the "nonlocal" samples for each class is itself a nontrivial problem. Note that here our goal is face classification but not face reconstruction, and using the Euclidian/cosine distance to identify the nonlocal samples may not be effective. Second, by introducing the nonlocal samples to represent the query sample, all the classes will have a smaller representation residual, and thus the discrimination of representation residual can be reduced, making the classification harder. Third, such a strategy can be computationally expensive because for each class we need to identify the nonlocal samples and calculate the representation of the test sample.

In SRC [16] this "lack of samples" problem is solved by using the collaborative representation strategy, i.e., coding the query image $y$ over the samples from all classes $X = [X_1, X_2, ..., X_K]$ as $y \approx X\alpha$. Such a collaborative representation strategy simply takes the face images from all the other classes as the nonlocal samples of one class. Though this strategy is very simple, there are two points need to be stressed. First, the searching for nonlocal samples of each class can be avoided. Second, all the classes share one common representation of the query sample, and thus the conventional representation residual based classification procedure used in NN and NS classifiers cannot be used.

Though we call the representation of $y$ by $X$ "collaborative representation", we have no objection if anyone would like to call it "competitive representation", because each class will contribute competitively to represent $y$. If one class contributes more, this means that other classes will contribute less. In this face representation problem, "collaboration" and "competition" are the two sides of the same coin. Therefore, one intuitive but very effective classification rule is to check which class contributes the most in the collaborative representation of $y$, or equivalently which class has the least reconstruction residual. We call this classification scheme the collaborative representation based classification (CRC).

### 3.1. Discussions on collaborative representation based classification

After collaboratively represent $y$ using Eq. (1), SRC classifies $y$ by checking the representation residual class by class using Eqs. (2) and (3). To simplify the analysis, let's remove the $l_1$-regularization term in Eq. (1), and the representation becomes the least square problem: $\{\hat{\alpha}_i\} = \min_{\{\alpha_i\}} \|y - \sum_i X_i \alpha_i\|_2^2$. Refer to Fig. 3, the resolved representation $\hat{y} = \sum_i X_i \hat{\alpha}_i$ is the perpendicular projection of $y$ onto the space spanned by $X$. The reconstruction residual by each class is $r_i = \|y - X_i \hat{\alpha}_i\|_2^2$. It can be readily derived that

$$r_i = \|y - X_i \hat{\alpha}_i\|_2^2 = \|y - \hat{y}\|_2^2 + \|\hat{y} - X_i \hat{\alpha}_i\|_2^2$$

Obviously, when we use $r_i$ to determine the identity of $y$, it is the amount

$$r_i^* = \|\hat{y} - X_i \hat{\alpha}_i\|_2^2 \tag{7}$$

that works for classification because $\|y - \hat{y}\|_2^2$ is a constant for all classes.

From a geometric viewpoint, we can write $r_i^*$ as

$$r_i^* = \frac{\sin^2(\hat{y}, \chi_i) \| \hat{y} \|_2^2}{\sin^2(\chi_i, \bar{\chi}_i)} \quad (8)$$

where $\chi_i = X_i \hat{\alpha}_i$ is a vector in the space spanned by $X_i$, and $\bar{\chi}_i = \sum_{j \neq i} X_j \hat{\alpha}_j$ is a vector in the space spanned by all the other classes $X_j$, $j \neq i$. Eq. (8) shows that by using CRC, when we judge if $y$ belongs to class $i$, we will not only consider if the angle between $\hat{y}$ and $\chi_i$ is small (i.e., if $\sin(\hat{y}, \chi_i)$ is small), but also consider if the angle between $\chi_i$ and $\bar{\chi}_i$ is big (i.e., if $\sin(\chi_i, \bar{\chi}_i)$ is big). Such a "double checking" mechanism makes the CRC effective and robust for classification.

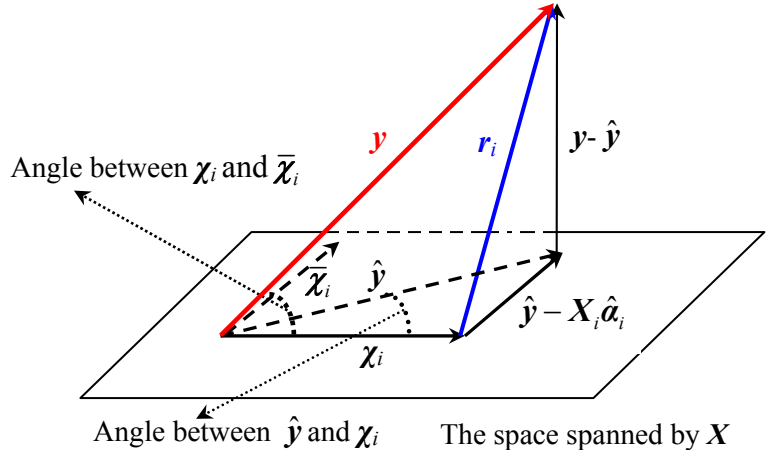

**Figure 3:** Illustration of collaborative representation based classification.

When the number of classes is too big, the number of atoms in dictionary $X = [X_1, X_2, \ldots, X_K]$ will be big so that the least square solution $\{\hat{\alpha}_i\} = \min_{\{\alpha_i\}} \| y - \sum_i X_i \alpha_i \|_2^2$ can become unstable. This problem can be solved by regularization. In SRC, the $l_1$-sparsity constraint is imposed on $\alpha$ to regularize the solution. However, the $l_1$-minimization is time consuming. As we will see in the section of experimental results, by using $l_2$-norm to regularize the solution of $\alpha$, we can have similar FR results to those by $l_1$- regularization but with significantly less complexity, implying that the collaborative representation plays a more important role than the $l_1$-norm regularization in the application of FR.

**3.2. General model of collaborative representation**

The coding of a query image $y$ over the dictionary $X$ can be written as $y = x + e$, where $x \approx X\alpha$ is the component we want to recover from *y* for classification use and *e* is the residual (e.g., noise, occlusion and corruption). A general model of collaborative representation is:

$$\hat{\boldsymbol{\alpha}} = \arg\min_{\boldsymbol{\alpha}} \left\{ \|\boldsymbol{y} - \boldsymbol{X}\boldsymbol{\alpha}\|_{l_q} + \lambda \|\boldsymbol{\alpha}\|_{l_p} \right\} \qquad (9)$$

where $\lambda$ is the regularization parameter and *p*, *q* = 1 or 2. Different settings of *p* and *q* lead to different instantiations of the collaborative representation model. For example, in SRC [16] *p* is set as 1 while *q* is set as 1 or 2 to handle face recognition with and without occlusion/corruption, respectively.

Different from image restoration, where the goal is to faithfully reconstruct the signal from the noisy and/or incomplete observation, in CRC the goal is twofold. First, we want to recover the desired signal *x* from *y* with the resolved coding vector $\hat{\boldsymbol{\alpha}}$ (i.e., *x*=*X*$\hat{\boldsymbol{\alpha}}$) such that in *x* the noise and trivial information can be suppressed. Second, in order for an accurate classification, the coding vector $\hat{\boldsymbol{\alpha}}$ should be sparse enough so that the identity of *y* can be easily identified. The question is how to set *p* and *q* in Eq. (9) to achieve the above goals with a reasonable degree of computational complexity.

In the case that there is no occlusion/corruption in *y* (the case that *y* is occluded/corrupted will be discussed in sub-section 3.4), we may assume that the observed image *y* contains some additive Gaussian noise. Under such an assumption, it is known that the $l_2$-norm should be used to characterize the data fidelity term in Eq. (9) in order for a *maximum a posterior* (MAP) estimation of *x* [13]. Thus we set *q*=2. Let's then discuss the regularization term in Eq. (9). Most of the previous works [16-19] such as SRC impose $l_1$-regularization on $\boldsymbol{\alpha}$, and it is believed that the $l_1$-regularization makes the coding vector $\hat{\boldsymbol{\alpha}}$ sparse. In order to investigate the role of $l_1$-regularization on $\boldsymbol{\alpha}$, let's conduct some experiments to plot the distribution of $\hat{\boldsymbol{\alpha}}$.

We use the Extended Yale B and AR databases to perform the experiments. The training samples (1216 samples in Extended Yale B and 700 samples in AR) are used as the dictionary *X*. The PCA is used to reduce the dimensionality of face images. For each test sample *y*, it is coded over *X*, and the coding vector calculated from all the test samples are used to draw the histogram of $\hat{\boldsymbol{\alpha}}$. In the first experiment, we reduce the dimensionality of face images to 800 for Extended Yale B and 500 for AR. Then the dictionaries *X* for the two databases are of size 800×1216 and 500×700, respectively. Since both the two systems are under-determined, we calculate the coding vector by least-square method but with a weak regularization:

$\hat{\boldsymbol{\alpha}} = \left( \boldsymbol{X}^T \boldsymbol{X} + 0.0001 \cdot \boldsymbol{I} \right)^{-1} \boldsymbol{X}^T \boldsymbol{y}$. In Figs. 4(a) and 4(b) we draw the histograms of $\hat{\boldsymbol{\alpha}}$ for the two databases, as well as the fitted curves of them by using Gaussian and Laplacian functions.

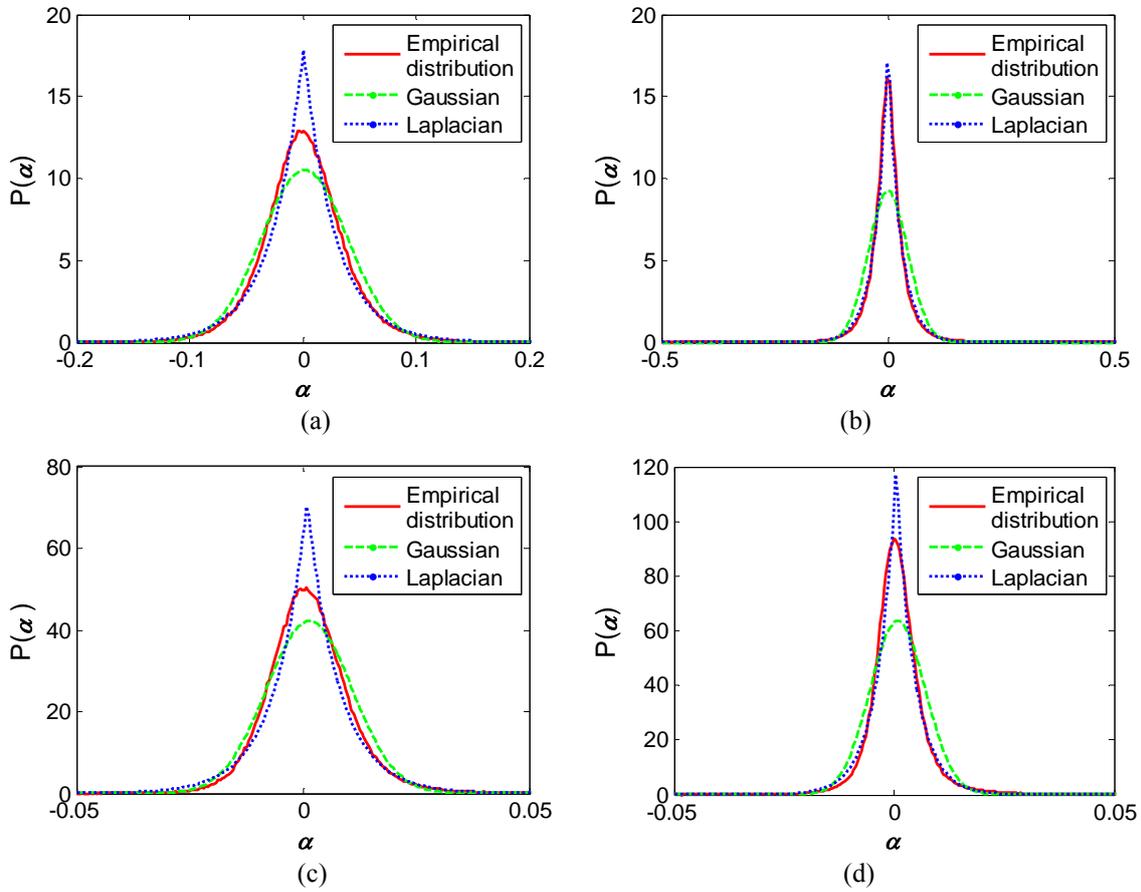

**Figure 4:** The histograms (in red) of the coding coefficients and the fitted curves of them by using Gaussian (in green) and Laplacian (in blue) functions. (a) and (b) show the curves for AR (dimension: 500) and Extended Yale B (dimension: 800) databases, respectively, while (c) and (d) show the curves when the feature dimension is 50.

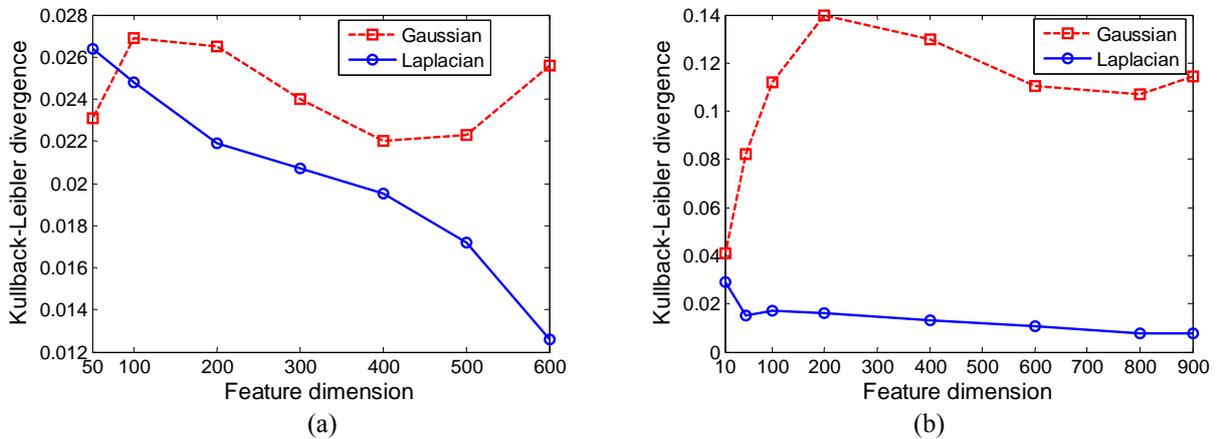

**Figure 5:** The Kullback-Leibler divergences between the coding coefficient histograms and the fitted curves (by Gaussian and Laplacian distributions) under different feature dimensions. (a) AR; and (b) Extended Yale B.

From Figs. 4(a) and 4(b), we can see that the distribution of $\hat{\alpha}$ can be much better fitted as Laplacian than Gaussian. The Kullback-Leibler divergences between the histograms and the fitted curves are 0.0223 by Gaussian and 0.0172 by Laplacian for the AR database, and 0.1071 by Gaussian and 0.0076 by Laplacian for the Extended Yale B database. It can be seen that via collaborative representation the distribution of $\hat{\alpha}$ naturally and *passively* tends to be sparse (i.e., Laplacian) even without $l_1$-regularization. This is because when the dimension of face feature $y$ is relatively high (e.g., 500), usually the discrimination of $y$ is also high so that only a few training samples, mostly from the same class as $y$, are involved to code it. This leads to a natural sparse representation of $y$.

We then reduce the face feature dimensionality to 50 by PCA, and draw in Figs. 4(c) and 4(d) the histograms of $\hat{\alpha}$ as well as the fitted curves of them. It can be found that the accuracy of Laplacian fitting is reduced (the Kullback-Leibler divergences are 0.0264 for the AR database and 0.0152 for the Extended Yale B database), while the Gaussian fitting of the histogram is much improved (the Kullback-Leibler divergences are 0.0231 for the AR database and 0.0820 for the Extended Yale B database). This is because when the dimension of the face feature $y$ is low (e.g., 50), the discrimination of $y$ will be much decreased so that quite a few training samples from various classes will be involved to code $y$. This makes the representation of $y$ much less sparse, and raises the difficulty to correctly identify $y$.

For a more comprehensive observation of the relationship between the dimensionality of $y$ and the sparsity of coding coefficient $\alpha$, in Fig. 5 we show the Kullback-Leibler divergences between the coding coefficient histograms and the fitted Gaussian and Laplacian functions under various feature dimensions. Clearly, with the increase of feature dimensionality, the fitting error by Laplacian function decreases, implying that the increase of feature discrimination can naturally force the coding coefficients to be sparsely distributed. In such case, it is not necessary to further regularize $\alpha$ by using the expensive $l_1$-norm regularization. However, with the decrease of feature dimensionality, the discrimination power of the feature vector will also decrease, and the distribution of $\alpha$ becomes less sparse. In this case, we may need to impose $l_1$-regularization on $\alpha$ to *actively* sparsify $\alpha$ and thus enhance the classification accuracy. Our experiments in Section 4 will also validate the above analyses.

### 3.3. CRC with regularized least square

In practical FR systems, usually the feature dimensionality will not be set too low in order for a good recognition rate. Therefore, we may not need to use $l_1$-regularization to sparsify $\alpha$. Considering that the dictionary $X$ can be under-determined, we use $\|\alpha\|_2$ to regularize the solution of Eq. (9), leading to the following regularized least square (RLS) instantiation of collaborative representation:

$$\hat{\alpha} = \arg\min_{\alpha} \left\{ \|y - X\alpha\|_2^2 + \lambda \|\alpha\|_2^2 \right\} \tag{10}$$

The role of $l_2$-regularization term $\|\alpha\|_2$ is two-folds. First, it makes the least square solution stable, particularly when $X$ is under-determined; second, it introduces a certain amount of sparsity to $\hat{\alpha}$, yet this sparsity is much weaker than that by $l_1$-regularization.

The solution of RLS based collaborative representation in Eq. (10) can be analytically derived as $\hat{\alpha} = Py$, where $P = \left(X^T X + \lambda \cdot I\right)^{-1} X^T$. Clearly, $P$ is independent of $y$ so that it can be pre-calculated. Once a query sample $y$ comes, we can simply project $y$ onto $P$ via $Py$. This makes the coding very fast. The classification by $\hat{\alpha}$ is similar to that in SRC (refer to Table 1 please). In addition to use the class-specified representation residual $\|y - X_i \hat{\alpha}_i\|_2$ for classification, where $\hat{\alpha}_i$ is the coding vector associated with class $i$, the $l_2$-norm "sparsity" $\|\hat{\alpha}_i\|_2$ also brings some discrimination information. We propose to use both of them in the decision making. (Based on our experiments, this improves slightly the classification accuracy over that by using only $\|y - X_i \hat{\alpha}_i\|_2$.) The proposed CRC algorithm via RLS (CRC-RLS) is summarized in Table 2.

**Table 2**: The CRC-RLS Algorithm

1. Normalize the columns of $X$ to have unit $l_2$-norm.
2. Code $y$ over $X$ by
$$\hat{\alpha} = Py \tag{11}$$
where $P = \left(X^T X + \lambda \cdot I\right)^{-1} X^T$.
3. Compute the regularized residuals
$$r_i = \|y - X_i \hat{\alpha}_i\|_2 / \|\hat{\alpha}_i\|_2 \tag{12}$$
4. Output the identity of $y$ as
$$\text{identity}(y) = \arg\min_i \{r_i\} \tag{13}$$

When a new subject is enrolled, the dictionary $X$ should be updated as $X = [X_o\ X_n]$, where $X_o$ is the original training data matrix and $X_n$ is composed of the training samples of the new subject. The projection matrix $P$ should also be recomputed as $P = \left(X^T X + \lambda \cdot I\right)^{-1} X^T$ with the updated dictionary $X$.

**3.4. Robust CRC (R-CRC) to occlusion/corruption**

In Section 3.3, we discussed FR without face occlusion/corruption and used $l_2$-norm to model the coding residual. However, when there are outliers (e.g., occlusions and corruptions) in the query face images, using $l_1$-norm to measure the representation fidelity is more robust than $l_2$-norm because $l_1$-norm could tolerate the outliers. In the robust version of SRC (R-SRC), the $l_1$-norm is used to measure the coding residual for robustness to occlusions/corruptions. In CRC, we could also adopt $l_1$-norm to measure the coding residual, leading to the robust CRC (R-CRC) model:

$$\hat{\alpha} = \arg\min_{\alpha} \left\{\|y - X\alpha\|_1 + \lambda \|\alpha\|_2^2\right\} \tag{14}$$

Let $e = y - X\alpha$. Eq. (14) can be re-written as

$$\hat{\alpha} = \arg\min_{\alpha} \left\{\|e\|_1 + \lambda \|\alpha\|_2^2\right\} \text{ s.t. } y = X\alpha + e \tag{15}$$

Eq. (15) is a constrained convex optimization problem which can be efficiently solved by the Augmented Lagrange Multiplier (ALM) method [55, 56]. The corresponding augmented Lagrangian function is

$$L_\mu(e, \alpha, z) = \|e\|_1 + \lambda \|\alpha\|_2^2 + \langle z, y - X\alpha - e\rangle + \frac{\mu}{2}\|y - X\alpha - e\|_2^2 \tag{16}$$

where $\mu > 0$ is a constant that determines the penalty for large representation error, and $z$ is a vector of Lagrange multiplier. The ALM algorithm iteratively estimates the Lagrange multiplier and the optimal solution by iteratively minimizing the augmented Lagrangian function

$$(e_{k+1}, \alpha_{k+1}) = \arg\min_{e,\alpha} L_{\mu_k}(e, \alpha, z_k) \tag{17}$$

$$z_{k+1} = z_k + \mu_k(y - X\alpha - e) \tag{18}$$

The above iteration converges to the optimal solution of Eq. (15) when $\{\mu_k\}$ is a monotonically increasing positive sequence [55].

The minimization in the first stage (i.e., Eq. (17)) of the ALM iteration could be implemented by alternatively and iteratively updating the two unknowns $e$ and $\alpha$ as follows:

$$\begin{cases} \boldsymbol{\alpha}_{k+1} = \arg\min_{\boldsymbol{\alpha}} L_{\mu_k}(\boldsymbol{\alpha}, \boldsymbol{e}_k, \boldsymbol{z}_k) \\ \boldsymbol{e}_{k+1} = \arg\min_{\boldsymbol{e}} L_{\mu_k}(\boldsymbol{\alpha}_{k+1}, \boldsymbol{e}, \boldsymbol{z}_k) \end{cases} \quad (19)$$

for which we could have a closed-form solution:

$$\begin{cases} \boldsymbol{\alpha}_{k+1} = \left(\boldsymbol{X}^T \boldsymbol{X} + 2\lambda/\mu_k \boldsymbol{I}\right)^{-1} \boldsymbol{X}^T \left(\boldsymbol{y} - \boldsymbol{e}_k + \boldsymbol{z}_k/\mu_k\right) \\ \boldsymbol{e}_{k+1} = S_{1/\mu_k}\left[\boldsymbol{y} - \boldsymbol{X}\boldsymbol{\alpha}_{k+1} + \boldsymbol{z}_k/\mu_k\right] \end{cases} \quad (20)$$

where the function $S_\alpha$, $\alpha \geq 0$, is the shrinkage operator defined component-wise as

$$[S_\alpha(\boldsymbol{x})]_i = \text{sign}(x_i) \cdot \max\{|x_i| - \alpha, 0\} \quad (21)$$

Clearly, $\boldsymbol{P}_k = \left(\boldsymbol{X}^T \boldsymbol{X} + 2\lambda/\mu_k \boldsymbol{I}\right)^{-1} \boldsymbol{X}^T$ is independent of $\boldsymbol{y}$ for the given $\mu_k$ and thus $\{\boldsymbol{P}_k\}$ can be pre-calculated as a set of projection matrices. Once a query sample $\boldsymbol{y}$ comes, in the first stage of ALM we can simply project $\boldsymbol{y}$ onto $\boldsymbol{P}_k$ via $\boldsymbol{P}_k\boldsymbol{y}$. This makes the calculation very fast. After solving the representation coefficients $\boldsymbol{\alpha}$ and residual $\boldsymbol{e}$, similar classification strategy to CRC-RLS can be adopted by R-CRC. The entire algorithm of R-CRC is summarized in Table 3.

**Table 3**: The R-CRC Algorithm

1. Normalize the columns of $\boldsymbol{X}$ to have unit $l_2$-norm.
2. Code $\boldsymbol{y}$ over $\boldsymbol{X}$ by
   **INPUT:** $\boldsymbol{\alpha}_0$, $\boldsymbol{e}_0$ and $\tau > 0$.
   **WHILE** not converged **Do**
   $$\boldsymbol{\alpha}_{k+1} = \left(\boldsymbol{X}^T \boldsymbol{X} + 2\lambda/\mu_k \boldsymbol{I}\right)^{-1} \boldsymbol{X}^T \left(\boldsymbol{y} - \boldsymbol{e}_k + \boldsymbol{z}_k/\mu_k\right)$$
   $$\boldsymbol{e}_{k+1} = S_{1/\mu_k}\left[\boldsymbol{y} - \boldsymbol{X}\boldsymbol{\alpha}_{k+1} + \boldsymbol{z}_k/\mu_k\right]$$
   $$\boldsymbol{z}_{k+1} = \boldsymbol{z}_k + \mu_k\left(\boldsymbol{y} - \boldsymbol{X}\boldsymbol{\alpha}_{k+1} - \boldsymbol{e}_{k+1}\right)$$
   **End WHILE**
   **OUTPUT:** $\hat{\boldsymbol{\alpha}}$ and $\hat{\boldsymbol{e}}$.
3. Compute the regularized residuals
   $$r_i = \|\boldsymbol{y} - \boldsymbol{X}_i\hat{\boldsymbol{\alpha}}_i - \hat{\boldsymbol{e}}\|_2 / \|\hat{\boldsymbol{\alpha}}_i\|_2$$
4. Output the identity of $\boldsymbol{y}$ as
   $$\text{identity}(\boldsymbol{y}) = \arg\min_i \{r_i\}$$

## 4. Experimental Results

In this section, we conduct extensive experiments to evaluate the various instantiations of CRC. We would like to stress that the goal of this paper is not to argue that CRC has better FR accuracy than SRC, but rather to investigate the roles of sparsity and collaborative representation in FR. SRC itself is an instantiation of CRC by using $l_1$-norm to regularize the coding coefficients and/or to model the coding residual. As in [16], in the following experiments it is usually assumed that the face images are already aligned.

In Sections 4.1~4.4, considering the accuracy and computational efficiency we chose $l_1\_ls$ [22] to solve the $l_1$-regularized SRC scheme. In the experiment of gender classification, the parameter $\lambda$ of CRC-RLS and RNS_$l_p$ ($p$=1 or 2) is set as 0.08. In FR experiments, when more classes (and thus more samples) are used for collaborative representation the least square solution will become more unstable and thus higher regularization is required. We set $\lambda$ as 0.001·$n$/700 for CRC-RLS in all FR experiments, where $n$ is the number of training samples. If no specific instruction, for R-CRC we set $\lambda$ as 1 in FR with occlusion.

Five benchmark face databases, the Extended Yale B [46] [50], AR [57], Multi-PIE [58], LFW [63] and FRGC version 2.0 [61], are used in evaluating CRC and its competing methods, including SRC, SVM, LRC [48], and NN. (Note that LRC is an NS based method.) All the experiments were implemented using MATLAB on a 3.16 GHz machine with 3.25GB RAM. In Section 4.1, we use examples to discuss the role of $l_1$-norm and $l_2$-norm regularizations; in Section 4.2, we use gender classification as an example to illustrate that collaborative representation is not necessary when there are enough training samples of each class; FR without and with occlusion/corruption are conducted in Section 4.3 and Section 4.4, respectively; face validation is conducted in Section 4.5; finally the running time of SRC and CRC is evaluated in Section 4.6.

### 4.1. $L_1$-regularization vs. $L_2$-regularization

In this section, we study the role of sparse regularization in FR by using the Extended Yale B [46][50] and AR [57] databases (the experimental setting will be described in Section 4.3). The Eigenfaces with dimensionality 300 are used as the input facial features. The dictionary is formed by all the training samples.

We test the performance of S-SRC ($l_1$-regularized minimization) and CRC-RLS ($l_2$-regularized minimization) with different values of regularization parameter $\lambda$ in Eq. (1) and Eq. (10). The results on the AR and Extended Yale B databases are shown in Fig. 6(a) and Fig. 6(b), respectively. We can see that when $\lambda$=0, both S-SRC and CRC-RLS will fail. When $\lambda$ is assigned a small positive value, e.g., from 0.000001 to 0.1, good results can be achieved by S-SRC and CRC-RLS. When $\lambda$ is too big (e.g., >0.1) the recognition rates of both methods fall down. With the increase of $\lambda$ (>0.000001), no much benefit on recognition rate can be gained. In addition, the $l_2$-regularized minimization (i.e., CRC-RLS) could get similar recognition rates to the $l_1$-regularized minimization (i.e., S-SRC) in a broad range of $\lambda$. This validates our discussion in Section 3.2 that the $l_1$-regularization on $\alpha$ is not necessary when the discrimination of face feature is high enough. However, when the dimension of facial features is very low, the representation will become very underdetermined, and the FR results by $l_1$-norm and $l_2$-norm regularizations could be substantially different, as demonstrated in [60] and discussed in Section 3.2 of this paper. In such case, $l_1$-regularization is helpful to get discriminative coefficients for accurate FR.

Fig. 6(c) plots one query sample's coding coefficients by S-SRC and CRC-RLS when they achieve their best results in the AR database. It can be seen that CRC-RLS has much weaker sparsity than S-SRC; however, it achieves no worse FR results. Again, $l_1$-sparsity is not crucial for FR when the facial feature is discriminative, while the collaborative representation mechanism in CRC-RLS and S-SRC is very helpful.

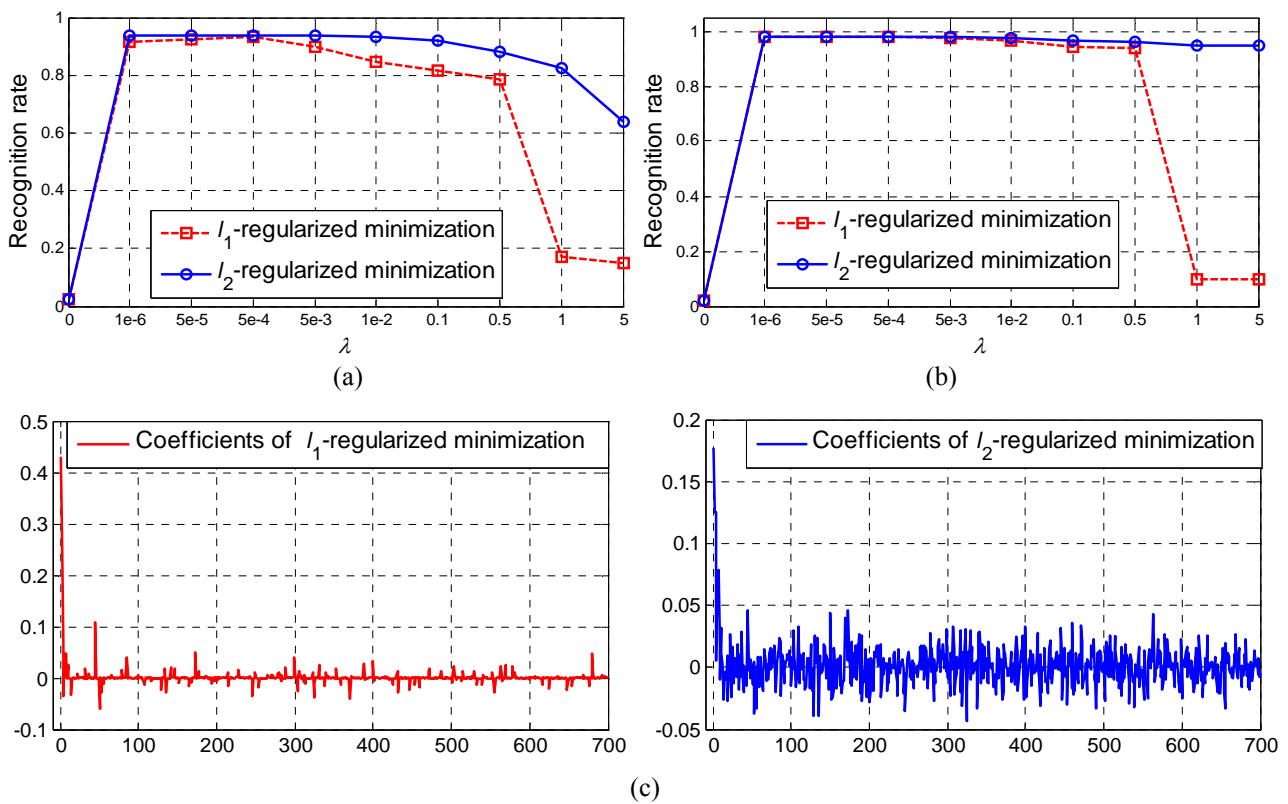

**Figure 6:** The recognition rates of S-SRC ($l_1$-regularized minimization) and CRC-RLS ($l_2$-regularized minimization) versus the different values of $\lambda$ on the (a) AR and (b) Extended Yale B databases. The coding coefficients of one query sample are plotted in (c).

## 4.2. Gender classification

In Section 2, we indicate that when each class has enough number of samples, there is no need to code the query sample over all classes because the subset of each class can form a nearly over-complete dictionary already. To validate this claim, we conduct experiments on a two-class separation problem: gender classification. We chose a non-occluded subset (14 images per subject) of AR [57], which consists of 50 male and 50 female subjects. Images of the first 25 males and 25 females were used for training, and the remaining images for testing. PCA is used to reduce the dimension of each image to 300. Since there are enough training samples in each class, as we discussed in Section 2, the RNS_$l_p$ (please refer to Eq. (6) and the related explanations) methods could do a good job for this gender classification task.

We compare RNS_$l_1$ and RNS_$l_2$ with the CRC-RLS, S-SRC, SVM, LRC, and NN methods. The results are listed in Table 4. One can see that RNS_$l_1$ and RNS_$l_2$ get the same best results, validating that coding on each class is more effective than coding on all classes when the training samples per class are sufficient, no matter $l_1$- or $l_2$-regularizaion is used. CRC-RLS gets the second best result, about 1.4% higher than S-SRC. The nearest subspace method (e.g., LRC) achieves much worse results because wrong class may have lower representation residual than correct class without regularization on the coding coefficient.

By using the above AR dataset as the training set, we then conducted cross-database gender classification by using a subset of the Multi-PIE database [58] as the test set. The face images of the first 250 subjects (including 174 males and 76 females) in Sessions 2, 3, and 4 were employed as the test images. In each session, each subject has 10 frontal face images with even number illuminations. The experimental results are shown in Table 5. It can be seen that all methods have lower classification rates than those in Table 4 because cross-dataset gender classification is more challenging due to the different data collection settings and environments. Nevertheless, RNS_$l_2$ achieves the best results in all sessions. RNS_$l_1$ and RNS_$l_2$ work better than S-SRC and CRC-RLS, respectively, which again validates that collaborative representation is not necessary when there are sufficient training samples of each class.

**Table 4:** The results of different methods on gender classification using the AR database.

| RNS_$l_1$ | RNS_$l_2$ | CRC-RLS | S-SRC | SVM | LRC | NN |
|---|---|---|---|---|---|---|
| **94.9%** | **94.9%** | 93.7% | 92.3% | 92.4% | 27.3% | 90.7% |

**Table 5:** The results of different methods on gender classification across the MPIE and AR databases.

| Session | RNS_$l_1$ | RNS_$l_2$ | CRC-RLS | S-SRC | SVM | LRC | NN |
|---|---|---|---|---|---|---|---|
| Session 2 | 77.3% | **79.6%** | 78.4% | 77.2% | 77.9% | 28.9% | 70.2% |
| Session 3 | 78.0% | **80.0%** | 79.3% | 77.9% | 76.6% | 31.9% | 70.6% |
| Session 4 | 79.0% | **80.7%** | 79.5% | 77.9%; | 78.3% | 29.7% | 70.3% |

## 4.3. Face recognition without occlusion/corruption

We then test the proposed CRC-RLS method for FR without occlusion/corruption. The Eigenface is used as the face feature in these experiments.

*1) Extended Yale B Database:* The Extended Yale B [46][50] database contains about 2,414 frontal face images of 38 individuals. We used the cropped and normalized face images of size 54×48, which were taken under varying illumination conditions. We randomly split the database into two halves. One half, which contains 32 images for each person, was used as the dictionary, and the other half was used for testing. Table 6 shows the recognition rates versus feature dimension by NN, LRC, SVM, S-SRC and CRC-RLS. Considering that each class has a good number (about 32) of training samples, here we also report the performance of RNS_$l_2$. It can be seen that CRC-RLS and S-SRC achieve very similar recognition rates. When the feature dimensionality is relatively high (e.g., 150 and 300), the difference of their recognition rates is less than 0.5%. When the feature dimensionality is set very low (e.g., 50), S-SRC will show certain advantage over CRC-RLS in terms of recognition rate. This is exactly in accordance with our analysis in Section 3.2. RNS_$l_2$ has worse performance than CRC_RLS and SRC although it performs well with 50-dimension feature. Since in this experiment there are enough training samples per subject, the recognition rates of all methods are not bad.

**Table 6**: The face recognition results of different methods on the Extended Yale B database.

| Dim | 50 | 150 | 300 |
|---|---|---|---|
| NN | 78.5% | 90.0% | 91.6% |
| LRC | 93.1% | 95.1% | 95.9% |
| RNS_$l_2$ | **94.6%** | 95.8% | 96.3% |
| SVM | 93.4% | 96.4% | 97.0% |
| S-SRC | 93.8% | **96.8%** | **97.9%** |
| CRC-RLS | 92.5% | 96.3% | **97.9%** |

**Table 7**: The face recognition results of different methods on the AR database.

| Dim | 54 | 120 | 300 |
|---|---|---|---|
| NN | 68.0% | 70.1% | 71.3% |
| LRC | 71.0% | 75.4% | 76.0% |
| SVM | 69.4% | 74.5% | 75.4% |
| S-SRC | **83.3%** | 89.5% | 93.3% |
| CRC-RLS | 80.5% | **90.0%** | **93.7%** |

*2) AR database:* As in [8], the cropped AR dataset [62][57] (with only illumination and expression changes) that contains 50 male subjects and 50 female subjects was used in our experiments. For each subject, the seven images from Session 1 were used for training, with other seven images from Session 2 for testing. The images were cropped to 60×43. The comparison of competing methods is given in Table 7. We can see that CRC-RLS achieves the best result when the dimensionality is 120 or 300, while it is slightly worse than S-SRC when the dimensionality is very low (e.g., 54). This is again in accordance with our analysis in Section 3.2. The recognition rates of CRC-RLS and S-SRC are both at least 10% higher than other methods. This shows that collaborative representation do improve much face classification accuracy.

*3) Multi PIE database:* The CMU Multi-PIE database [58] contains images of 337 subjects captured in four sessions with simultaneous variations in pose, expression, and illumination. Among these 337 subjects, all the 249 subjects in Session 1 were used. For the training set, we used the 14 frontal images with 14 illuminations [3] and neutral expression. For the test sets, 10 typical even-number frontal images of illuminations taken with neutral expressions from Session 2 to Session 4 were used. The dimensionality of Eigenface is 300. Table 8 lists the recognition rates in the three tests. The results validate that CRC-RLS and S-SRC are the best in accuracy, and they have at least 6% improvement over the other three methods.

**Table 8**: The face recognition results of different methods on the MPIE database.

|    | NN    | LRC   | SVM   | S-SRC  | CRC-RLS |
|----|-------|-------|-------|--------|---------|
| S2 | 86.4% | 87.1% | 85.2% | 93.9%  | **94.1%** |
| S3 | 78.8% | 81.9% | 78.1% | **90.0%** | 89.3%  |
| S4 | 82.3% | 84.3% | 82.1% | **94.0%** | 93.3%  |

*4) FRGC database:* FRGC version 2.0 [61] is a large-scale face database established under uncontrolled indoor and outdoor settings. We use a subset (316 subjects having no less than 10 samples) of query face image database, which has large lighting and expression variations and image blur, etc. We randomly choose 9 images per subject as the training set (2844 image in total), with the remaining as the test set (4474 images in total). The images were cropped to 128×168. The recognition rates of competing methods under different feature dimensions are given in Table 9. CRC_RLS and SRC lead to much better performance than NN, LRC and SVM, whose recognition rates almost have little improvement with the increase of feature dimension. When the feature dimension is no less than 400, the performance of CRC-RLS is very close to SRC, and CRC-RLS outperforms SRC a little in the case of 700-dimension feature.

---
[3] Illuminations {0,1,3,4,6,7,8,11,13,14,16,17,18,19}.

**Table 9**: The face recognition results of different methods on the FRGC 2.0 database.

| Dim | 100 | 400 | 700 |
|---|---|---|---|
| NN | 70.9% | 73.6% | 73.5% |
| LRC | 83.3% | 83.2% | 83.1% |
| SVM | 85.2% | 85.4% | 85.7% |
| SRC | **91.2%** | **95.6%** | 95.1% |
| CRC-RLS | 85.2% | 95.2% | **96.4%** |

To more comprehensively compare CRC-RLS with S-SRC, we plot their recognition rates using higher-dimensional features on the Extended Yale B, AR and FRGC2.0 databases in Fig 7. One can see that the curves of recognition rate become flat as the feature dimension increases, implying that using very high dimensional feature does not improve the FR accuracy. It can also be seen that CRC-RLS achieves similar performance to S-SRC when the feature dimension is not too low. Especially, on the large scale FRGC 2.0 database, CRC-RLS consistently outperforms S-SRC when using 700 or higher dimensional features.

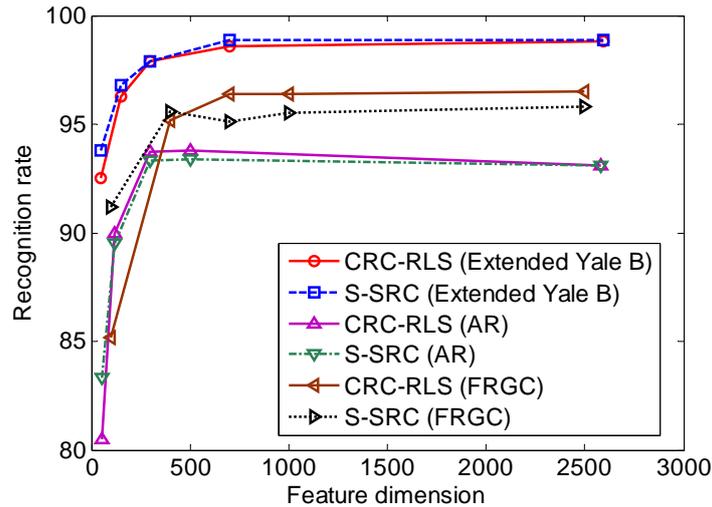

**Figure 7:** The comparison of CRC-RLC and S-SRC under different feature dimensions on different databases.

*5) LFW database:* Labeled Faces in the Wild (LFW) is a large-scale database designed for unconstrained FR with variations of pose, illumination, expression, misalignment and occlusion, etc. Two subsets of aligned LFW [63] are used here. Subset 1 consists of 311 subjects with no less than 6 samples per subject. We use the first 5 samples for training and the remaining samples for testing. Subset 2 consists of 143 subjects with no less than 11 samples per subject. We use the first 10 samples for training data and the remaining samples for testing. We divide a face image into 10×8 patches, and concatenate the histogram of Uniform-LBP [64] in each patch as the facial feature. The parameter $\lambda$ of CRC-RLS is set as 0.1. The recognition rates of competing methods are shown in Table 10. The recognition rates of all methods are much lower than those in the Extended Yale B, AR, Multi-PIE and FRGC databases because LFW database has much more uncontrolled variations (e.g.., pose and misalignment). CRC-RLS and S-SRC significantly outperform the other methods, while CRC-RLS performs slightly better than S-SRC.

**Table 10:** The face recognition results of different methods on the LFW database.

| Test | NN | LRC | SVM | S-SRC | CRC-RLS |
|---|---|---|---|---|---|
| LFW-6 | 30.2% | 33.8% | 44.3% | 53.5% | **54.0%** |
| LFW-11 | 45.9% | 52.9% | 63.0% | 75.5% | **76.8%** |

**4.4. Face recognition with occlusion/corruption**

One important advantage of representation (or coding) based FR methods is their ability to deal with occlusion and corruptions. In R-SRC [16], the robustness to face occlusion/corruption is achieved by adding an occlusion dictionary (an identity matrix) for sparse coding, or equivalently, using $l_1$-norm to measure the coding residual. In Section 3.4, we have correspondingly proposed the robust version of CRC, i.e., R-CRC, for FR with occlusion/corruption. In this subsection we test R-CRC in handling different kinds of occlusions, including random pixel corruption, random block occlusion and real disguise.

*1) FR with block occlusion:* To be identical to the experimental settings in [16], we used Subsets 1 and 2 (717 images, normal-to-moderate lighting conditions) of the Extended Yale B database for training, and used Subset 3 (453 images, more extreme lighting conditions) for testing. The images were resized to 96×84. As in [16], we simulate various levels of contiguous occlusion, from 0% to 50%, by replacing a randomly located square block of each test image with an unrelated image. The block occlusion of a certain size is located on the random position which is unknown to the FR algorithms. Here $\lambda$ of R-CRC is set as 0.1. The results by NN, LRC, S-SRC, R-SRC, CRC-RLS and R-CRC are shown in Table 11. We can see that R-CRC outperforms R-SRC in most cases (with 17% improvement in 50% occlusion) except for the case of 30% block occlusion. In addition, CRC-RLS achieves much better performance than S-SRC. This is mainly because the test sample with block occlusion cannot be well represented by the non-occluded training samples with sparse coefficients. In the following experiments, we only report the results of R-SRC in FR with corruption or disguise.

**Table 11:** The recognition rates of R-CRC, CRC-RLS, R-SRC and S-SRC under different levels of block occlusion.

| Occlusion | 0%   | 10%   | 20%   | 30%   | 40%   | 50%   |
|-----------|------|-------|-------|-------|-------|-------|
| NN        | 94.0%| 92.9% | 85.4% | 73.7% | 62.9% | 45.7% |
| LRC       | **100%** | **100%** | 95.8% | 81.0% | 63.8% | 44.8% |
| S-SRC     | **100%** | 99.6% | 93.4% | 77.5% | 60.9% | 45.9% |
| R-SRC     | **100%** | **100%** | 99.8% | **98.5%** | 90.3% | 65.3% |
| CRC-RLS   | **100%** | **100%** | 95.8% | 85.7% | 72.8% | 59.2% |
| R-CRC     | **100%** | **100%** | **100%** | 97.1% | **92.3%** | **82.3%** |

*2) FR with pixel corruption:* In this part, we test the robustness of R-SRC and R-CRC to pixel corruption. We used the same experimental settings as in [16], i.e., Subsets 1 and 2 of Extended Yale B for training and Subset 3 for testing. The images were resized to 96×84 pixels. For each test image, we replaced a certain percentage of its pixels by uniformly distributed random values within [0, 255]. The corrupted pixels were randomly chosen for each test image and the locations are unknown to the algorithm. Table 12 lists the recognition rates of NN, LRC, R-SRC, CRC-RLS and R-CRC. It can be seen that R-CRC achieves equal or better performance (about 13% improvement over R-SRC in 80% corruption) in almost all cases. Interestingly, CRC-RLS can also perform well up to 50% pixel corruption.

**Table 12:** The recognition rates of R-SRC, CRC-RLS and R-CRC under different levels of pixel corruption.

| Corruption | 0%   | 10%  | 20%  | 30%  | 40%  | 50%  | 60%  | 70%  | 80%  | 90%  |
|------------|------|------|------|------|------|------|------|------|------|------|
| NN         | 94.0%| 96.7%| 97.1%| 94.5%| 85.4%| 68.4%| 46.8%| 25.4%| 11.0%| 4.6% |
| LRC        | **100%** | **100%** | **100%** | 99.1%| 95.6%| 80.4%| 50.3%| 26.0%| 9.9% | 6.2% |
| R-SRC      | **100%** | **100%** | **100%** | **100%** | **100%** | **100%** | 99.3%| **90.7%** | 37.5%| 7.1% |
| CRC-RLS    | **100%** | **100%** | **100%** | 99.8%| 98.9%| 96.4%| 79.9%| 45.7%| 13.2%| 4.2% |
| **R-CRC**  | **100%** | **100%** | **100%** | **100%** | **100%** | **100%** | **100%** | 90.5%| **51.0%** | **15.9%** |

*3) FR with real face disguise:* As in [16], a subset from the cropped AR database [62] consisting of 1,200 images from 100 subjects, 50 male and 50 female, is used here. 800 images (about 8 samples per subject) of non-occluded frontal views with various facial expressions were used for training, while the others with sunglasses and scarves (as shown in Fig. 8) were used for testing. The images were resized to 83×60. The results of competing methods are shown in Table 13.

Although CRC-RLS is not designed for robust FR, interestingly it achieves the best result of FR with scarf disguise, outperforming SRC by a margin of 31% and R-CRC by 4.5%. (This phenomenon may result from the special experimental setting, which will be discussed more in the following FR experiments.) By using $l_1$-norm to measure the representation fidelity, R-CRC has the same recognition rate as R-SRC in sunglasses disguise, but achieves 26.5% improvement in scarf disguise. As in [16], we also partition the face image into 8 sub-regions for FR. With partition, CRC-RLS and R-CRC can still achieve slightly better performance than R-SRC in scarf disguise, but perform a little worse in sunglass disguise. This is because that for each partitioned face portion its discrimination is limited so that the $l_1$-regularization is helpful to improve the sparsity of coding vector and consequently improve the classification accuracy. Nevertheless, the recognition rates of CRC-RLS and R-CRC are very competitive with R-SRC.

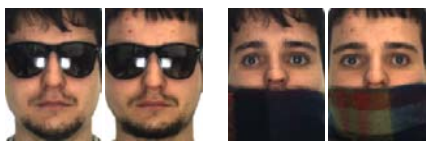

**Figure 8**: The test samples with sunglasses and scarves in the AR database.

**Table 13**: The results of different methods on face recognition with real disguise (AR database).

|  | Sunglass | Scarf |
| --- | --- | --- |
| R-SRC | **87.0%** | 59.5% |
| CRC-RLS | 68.5% | **90.5%** |
| R-CRC | **87.0%** | 86.0% |
| Partitioned | Sunglass | Scarf |
| R-SRC | **97.5%** | 93.5% |
| CRC-RLS | 91.5% | **95.0%** |
| R-CRC | 92.0% | **94.5%** |

**Table 14**: The results of face recognition with real disguise (AR database) by using intensity and LBP features.

| Intensity feature | Sunglass | Scarf |
| --- | --- | --- |
| R-SRC | **69.8%** | 40.8% |
| CRC-RLS | 57.2% | **71.8%** |
| R-CRC | 65.8% | **73.2%** |
| Histogram of LBP | Sunglass | Scarf |
| R-SRC | 92.5% | 94.8% |
| CRC-RLS | 93.5% | 95.0% |
| R-CRC | **94.2%** | **95.8%** |

In the above experiment of FR with scarf, the CRC-RLS model achieves higher recognition rates than the models with $l_1$-norm characterization of coding residual (i.e., R-SRC and R-CRC), while the reverse is true for FR with sunglasses. To have a more comprehensive observation of these methods' robustness to disguise, we perform another more challenging experiment. A subset from the AR database which consists of 1,900 images from 100 subjects, 50 male and 50 female, is used. 700 images (7 samples per subject) of non-occluded frontal views from session 1 were used for training, while all the images with sunglasses (or scarf) from the two sessions were used for testing (6 samples per subject per disguise). The images were resized to 83×60. Both the raw image intensity feature and the histogram of LBP [64] feature are used to evaluate the proposed method. The results are shown in Table 14. By using the image intensity feature, R-CRC is slightly worse than R-SRC in sunglass case with 4% gap, but significantly better than R-SRC in the scarf case with 32.4% improvement. Compared to R-SRC, CRC-RLS has 31% higher recognition rate in scarf case, and 13% lower rate in sunglass case. By using the local feature such as histogram of LBP, all methods could have much better performance (over 92% recognition accuracy), and both R-CRC and CRC-RLS outperform R-SRC. One can also see that R-CRC achieves better performance than CRC-RLS in these two cases, validating that $l_1$-regularization on representation residual is more robust than $l_2$-regulariation.

From Table 13 and Table 14, we may have the following findings. Since eyes are probably the most discriminative part in human face, the sunglass disguise will reduce a lot the discrimination of face image, and hence the $l_1$-regularized R-SRC method will show advantage in dealing with sunglass disguise because the $l_1$-regularization could actively increase the sparsity of coding coefficients. (Please refer to Section 3.2 for more discussions on the relationship between feature discrimination and coefficient sparsity.) In the case of scarf disguise, though the occlusion area is big, the discrimination of face image is actually not much decreased. Therefore, the $l_2$-regularized CRC-RLS and R-CRC methods can perform well. On the contrary, the $l_1$-regularization in R-SRC will prevent the use of enough samples to represent the occluded face image so that its recognition rate is lower than CRC-RLS and R-CRC. When more effective features, e.g., histogram of LBP, are used, $l_2$-regularized collaborative representation can show better performance than $l_1$-norm regularized sparse representation.

**4.5. Face validation**

In practical FR systems, it is important to reject invalid face images which have no template in the database [16]. In this section we test the face validation performance of the proposed method. The *Sparsity Concentration Index* (SCI) proposed in [16] is adopted to do face validation with the coding coefficient. The large-scale Multi-PIE face database is used in this experiment. The 100 subjects in Session 1 were used as the training set and the first 250 subjects in Session 2, 3 and 4 were used as customer images. Each training subject has 14 frontal images with neutral expression and illuminations {0,1,3,4,6,7,8,11,13,14,16,17,18,19}. For the test set, 10 typical frontal images of illuminations {0, 2, 4, 6, 8, 10, 12, 14, 16, 18} taken with neutral expressions were used. The face images were normalized and cropped to the size of 50×40.

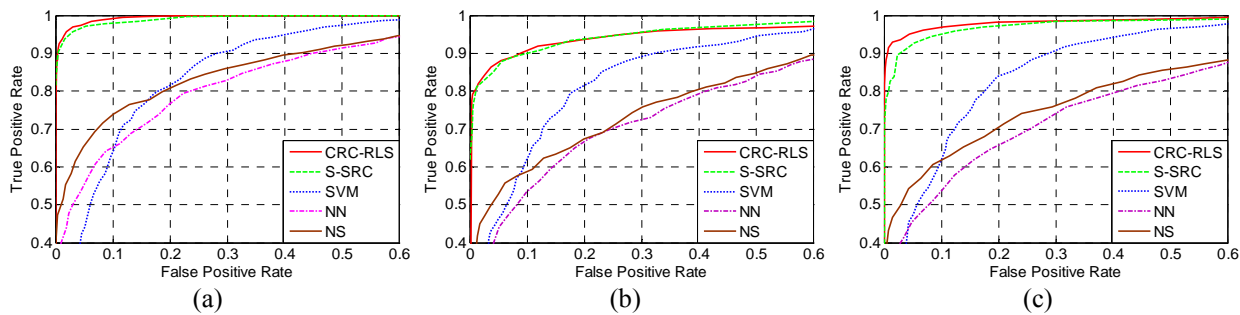

**Figure 9:** Subject validation on the MPIE database. (a) Session 2; (b) Session 3; and (c) Session 4.

Fig. 9 plots the ROC (receiver operating characteristic) curves of the competing methods: CRC-RLS, S-SRC, SVM, NN and NS. It can be seen that CRC-RLS and S-SRC work much better than the other methods, and CRC-RLS performs better than S-SRC in all sessions, especially Session 4. For instance, when the false positive rate is 0.1, the true positive rates of CRC-RLS are 99.1% in Session2, 90.7% in Session3 and 96.8% in Session4, while the true positive rates of S-SRC are 98.0%, 89.8% and 95.1% in the three Sessions, respectively. This is because the $l_1$-norm regularized sparse coding will force one specific class to represent the input invalid query sample, and hence incorrectly classify this sample to that class. Comparatively, $l_2$-norm regularized coding does not force the coding coefficients to be sparse, and allows the invalid query sample to be evenly represented across different classes. Therefore, the false recognition can be avoided.

### 4.6. Running time

We compare the running time of CRC and SRC under two situations. For FR without occlusion/corruption, it is good to use $l_2$-norm to measure the coding residual, and hence we compare the running time of S-SRC and CRC-RLS; for FR with occlusion/corruption, we compare the running time of R-SRC and R-CRC, where $l_1$-norm is used to measure the coding residual for robustness to outlier pixels.

*a) Face recognition without occlusion:* The running time of CRC-RLS and S-SRC with various fast $l_1$-minimization methods, including $l_1\_ls$ [36], ALM [34][31], FISTA [33] and Homotopy[35], are compared here. We fix the dimensionality of Eigenfaces as 300. The recognition rates and speed of S-SRC and CRC-RLS are listed in Table 15 (Extended Yale B), Table 16 (AR) and Table 17 (Multi-PIE), respectively. Note that the results in Table 17 are the average of Sessions 2, 3 and 4.

Table 15: Recognition rate and speed on the Extended Yale B database.

|  | Recognition rate | Time (s) |
|---|---|---|
| S-SRC($l_1$_ls) | **97.9%** | 5.3988 |
| S-SRC(ALM) | **97.9%** | 0.1280 |
| S-SRC(FISTA) | 91.4% | 0.1567 |
| S-SRC(Homotopy) | 94.5% | 0.0279 |
| CRC-RLS | **97.9%** | **0.0033** |
| **Speed-up** | **8.5 ~ 1636 times** | |

Table 16: Recognition rate and speed on the AR database.

|  | Recognition rate | Time (s) |
|---|---|---|
| S-SRC($l_1$_ls) | 93.3% | 1.7878 |
| S-SRC(ALM) | 93.3% | 0.0578 |
| S-SRC(FISTA) | 68.2% | 0.0457 |
| S-SRC(Homotopy) | 82.1% | 0.0305 |
| CRC-RLS | **93.7%** | **0.0024** |
| **Speed-up** | **12.6 ~ 744.9 times** | |

Table 17: Recognition rate and speed on the MPIE database.

|  | Recognition rate | Time (s) |
|---|---|---|
| S-SRC($l_1$_ls) | **92.6%** | 21.290 |
| S-SRC(ALM) | 92.0% | 1.7600 |
| S-SRC(FISTA) | 79.6% | 1.6360 |
| S-SRC(Homotopy) | 90.2% | 0.5277 |
| CRC-RLS | 92.2% | **0.0133** |
| **Speed-up** | **39.7 ~ 1600.7 times** | |

On the Extended Yale B database, CRC-RLS, S-SRC ($l_1$_ls) and S-SRC (ALM) achieve the best recognition rate (97.9%), but the speed of CRC-RLS is 1636 and 38.8 times faster than them. On the AR database, CRC-RLS has the best recognition rate and speed. S-SRC ($l_1$_ls) has the second best recognition rate but with the slowest speed. S-SRC (FISTA) and S-SRC (Homotopy) are much faster than S-SRC ($l_1$_ls) but they have lower recognition rates. On Multi-PIE, CRC-RLS achieves the second highest recognition rate (only 0.4% lower than S-SRC ($l_1$_ls)) but it is significantly (more than 1600 times) faster than S-SRC ($l_1$_ls). In this large-scale database, CRC-RLS is about 40 times faster than S-SRC with the fastest implementation (i.e., Homotopy), while achieving more than 2% improvement in recognition rate. We can see that the speed-up of CRC-RLS is more and more obvious as the scale (i.e., the number of classes or training samples) of face database increases, implying that it is more advantageous in practical large-scale FR applications.

*b) Face recognition with occlusion:* We compare the running time of R-CRC with R-SRC on the Multi-PIE corruption experiment [58]. As in [31] and [59], a subset of 249 subjects from Session 1 is used in this experiment. For each subject with frontal view, there are 20 images with different illuminations, among which the illuminations {0, 1, 7, 13, 14, 16, 18} were chosen as training images and the remaining 13 images were used as test data. The images were manually aligned and cropped to 40×30. For each test image, we replaced a certain percentage of its pixels by uniformly distributed random values within [0, 255]. The corrupted pixels were randomly chosen for each test image and the locations are unknown to the algorithm. The recognition rates and running time of R-SRC are directly copied from [31][59]. In order for a fair comparison of running time, we used a machine similar to that used in [31][59] to implement R-CRC[4].

**Table 18**: Average recognition rate with 50% and 70% random pixel corruptions on the MPIE database.

| Corruption | R-CRC | $l_1\_ls$ | Homotopy | SpaRSA | FISTA | ALM |
|---|---|---|---|---|---|---|
| 40% | **100%** | 97.8% | 99.9% | 98.8% | 99.0% | 99.9% |
| 50% | **100%** | 99.5% | 99.8% | 97.6% | 96.2% | 99.5% |
| 60% | 94.6% | 96.6% | **98.7%** | 90.5% | 86.8% | 96.2% |
| 70% | 68.4% | 76.3% | **84.6%** | 63.3% | 58.7% | 78.8% |

**Table 19**: The running time (second) of different methods versus corruption rate.

| Corruption | 0% | 20% | 40% | 60% | 80% | Average | **Speed-up** |
|---|---|---|---|---|---|---|---|
| $l_1\_ls$ | 19.48 | 18.44 | 17.47 | 16.99 | 14.37 | 17.35 | **18.94** |
| Homotopy | **0.33** | 2.01 | 4.99 | 12.26 | 20.68 | 8.05 | **8.79** |
| SpaRSA | 6.64 | 10.86 | 16.45 | 22.66 | 23.23 | 15.97 | **17.43** |
| FISTA | 8.78 | 8.77 | 8.77 | 8.80 | 8.66 | 8.76 | **9.56** |
| ALM | 18.91 | 18.85 | 18.91 | 12.21 | 11.21 | 16.02 | **17.49** |
| R-CRC | 0.916 | **0.914** | 0.918 | 0.916 | 0.915 | 0.916 | ----- |

Table 18 shows the FR rates of R-CRC and R-SRC implemented by various $l_1$-minimization solvers. One can see that R-CRC has the highest recognition rate in 40% and 50% corruption levels. In other cases, R-CRC is better than SpaRSA [32] and FISTA [33], and slightly worse than $l_1\_ls$ [36], Homotopy [35] and ALM [31]. The running time of different methods under various corruption levels is listed in Table 19. Apart from the case of 0% corruption, the proposed R-CRC has the lowest running time. It can also be seen that the running time of R-CRC is almost the same for all corruption levels. The speed-ups of R-CRC over R-SRC with various $l_1$-minimization algorithms are from 8.79 to 19.94 in average, showing that R-CRC has much lower time complexity.

## 5. Conclusions and Discussions

We discussed the role of $l_1$-norm regularization in the sparse representation based classification (SRC)

---

[4] Our MATLAB implementations are on a PC with dual quad-core 2.4G GHz Xeon processors and 16GB RAM, similar to that used in [31] and [59], in which the machine is with dual quad-core 2.66GHz Xeon processors and 8GB of memory.

scheme for face recognition (FR), and indicated that the collaborative representation nature of SRC plays a more important role than the $l_1$-regularization of coding vector in face representation and recognition. We then proposed a more general model, namely collaborative representation based classification (CRC). Two important instantiations of CRC, i.e., CRC via regularized least square (CRC-RLS) and robust CRC (R-CRC), were proposed for FR without and with occlusion/corruption, respectively. Compared with the $l_1$-regularized SRC, the $l_2$-regularized CRC-RLS and R-CRC have very competitive FR accuracy but with much lower time complexity, as demonstrated in our extensive experimental results.

SRC is also an instantiation of CRC by using $l_1$-norm to regularize the coding vector $\alpha$. The sparsity of $\alpha$ is related to the discrimination and dimension of face feature $y$. If the dimension is high, often the discrimination of $y$ is high and $\alpha$ will be naturally and *passively* sparse even without sparse regularization. In this case, $l_1$-regularization on $\alpha$ will not show advantage. If the dimension of $y$ is very low, often the discrimination of $y$ is low, and thus it is helpful to *actively* sparsify $\alpha$ by imposing $l_1$-regularization on it. In this case, using $l_1$-norm to regularize $\alpha$ will show visible advantage. In addition, in applications such as face validation, $l_1$-regularized coding will force one specific class to represent the invalid query sample and consequently recognize it (please see Section 4.5 for details). Comparatively, $l_2$-regularized coding does not force the coding vector to be sparse, and allows the invalid query sample to be evenly represented across different classes, and therefore avoids this problem.

Using $l_1$-norm to characterize the representation residual can be more tolerant to the outlier pixels in the query face image (e.g., occlusion, pixel corruption). However, this is effective only when the noises or errors in the query image are truly sparse or roughly follow the Laplacian distribution, which is not always true. For instance, in face image with real scarf, the distribution of the true error is not sparse and $l_2$-norm can even work better than $l_1$-norm to model the representation residual.